# FairSelect: A Systematic Evaluation of Multi-Level and Intersectional Algorithmic Fairness

FairSelect: Systematic, Multi-Level and Intersectional Fairness Evaluation

Nick Souligne*

College of Engineering, University of Arizona, Tucson, AZ, nasouligne@arizona.edu

Isabella Mixton-Garcia

College of Engineering, University of Arizona, Tucson, AZ, irmixtongarcia@arizona.edu

Vignesh Subbian

College of Engineering, University of Arizona, Tucson, AZ, vsubbian@arizona.edu

Algorithmic fairness methods are increasingly used to identify and mitigate bias in machine learning models, yet most approaches are evaluated in isolation and along single demographic axes. This limits practical guidance for selecting fairness strategies, where disparities may arise across intersectional subgroups and across multiple stages of the modeling lifecycle. This work presents *FairSelect*, a toolkit for systematically evaluating fairness mitigation strategies applied individually and in combination across pre-processing, in-processing, and post-processing stages. *FairSelect* supports multiple model architectures, intersectional subgroup evaluation, and comparison of fairness–utility tradeoffs across baseline, single-method, and multi-level configurations. The framework was validated using synthetic clinical datasets designed to represent specific bias mechanisms and a real-world replication of two-year stroke risk prediction among patients with atrial fibrillation. Synthetic experiments showed that targeted fairness methods generally reduced intended subgroup disparities, while combined strategies produced larger average fairness improvements with modest utility tradeoffs. In the clinical prediction task, mitigation effects were highly variable, with some combinations improving both fairness and predictive performance while others were ineffective or counterproductive. These findings demonstrate that fairness interventions interact in non-additive and context-dependent ways. *FairSelect* provides a practical framework for systematically identifying fairness strategies that improve subgroup equity while preserving model performance in clinical machine learning.

## 1 INTRODUCTION

The field of algorithmic fairness has produced an extensive and quickly expanding set of methods designed to identify and mitigate biases in machine learning models. These methods span multiple stages of the modeling lifecycle and often target distinct sources of biases, including pre-processing approaches that modify training data distributions, in-processing techniques that incorporate fairness constraints during model optimization, and post-processing strategies that adjust model outputs after training. Without these methods, the considerable promise of predictive modeling for improving patient

---

outcomes and healthcare delivery is fundamentally limited due to the biases embedded in the data, modeling approach, or arising through other sociotechnical processes that particularly impact historically marginalized populations [1-5]. While the application of algorithmic fairness approaches offers a mechanism to address these risks, their practical application remains opaque due to diversity of biases sources, non-generalizable methods, and a lack of a systemic approach to assess best practices.

Recent guidance from the STANDING Together collective has reinforced the need to move beyond overall model performance evaluation towards more explicit evaluation of subgroup performance with a focus on transparent reporting of mitigation strategies [6]. These recommendations call for users of health datasets to evaluate models in contextualized groups of interest, identify disparate performance in both these groups and others, and to report any methods used to modify performance across groups. Together, these motivate a need for subgroup-aware and intersectional fairness evaluation, but they do not specify how model developers should select among the many available mitigation techniques or determine whether combinations of methods improve or decrease fairness across the modeling lifecycle.

Different fairness techniques and measures are often designed to address distinct sources or manifestations of bias, such as representation imbalance, label bias, or decision threshold disparities [7]. This diversity suggests that applying multiple mitigation strategies in combination may improve overall fairness by targeting complementary sources of bias across the modeling pipeline. However, despite the abundance of available methods, fairness techniques are frequently developed, evaluated, and deployed in isolation, with most studies demonstrating the effectiveness of a single intervention within a narrowly defined experimental setting [1]. This separation reflects broader disciplinary and domain-specific silos, where fairness methods developed in one context may rely on assumptions about causal relationships, data structure, or outcome distribution that do not readily generalize to new application domains. As a result, model developers and practitioners are often left without clear guidance on how to select or combine fairness interventions, particularly when attempting to address multiple sources of bias simultaneously across different stages of the modeling lifecycle.

Recent work has begun to explore multi-level fairness strategies, defined as approaches that apply multiple mitigation techniques across one or more phases of the machine learning lifecycle [8,9,10]. Studies in domains such as finance and criminal justice have demonstrated that combining fairness interventions can produce meaningful reductions in bias while preserving predictive performance [11,12]. These studies are foundational to evaluating fairness mitigation strategies across multiple model types and intervention approaches. However, these prior frameworks have primarily focused on evaluating fairness along single protected attributes and as a result, little attention has been given to examining how combinations of fairness interventions affect intersectional subgroups, where disparities may emerge or persist at the intersections of multiple protected demographic characteristics.

Fairness assessments limited to single demographic attributes may underestimate disparities and fail to detect inequities affecting intersectional populations. However, incorporating intersectionality based approaches has been shown to improve model performance for these intersectional subgroups while maintaining high overall predictive performance [13,14,15]. Despite these findings, relatively little systematic work has been conducted to evaluate how combinations of fairness interventions perform across intersectional subgroups, including within multi-level fairness evaluation frameworks. This limitation is particularly relevant in healthcare contexts where intersectional disparities may have direct implications for clinical decision-making and patient outcomes.

Together, these limitations highlight the need for tools capable of systematically evaluating combinations of fairness interventions across multiple modeling stages while supporting intersectional fairness analysis. Importantly, fairness interventions should not be treated as independent components applied in isolation. Bias may arise at different points in the modeling pipeline, including through data representation, model learning steps, prediction errors, threshold selection, and downstream decision rules. Therefore, an intervention that reduces one form of disparity may not necessarily reduce another, and improvements in upstream fairness measures may not consistently improve downstream harm. Prior work has demonstrated this concern in showing that representational bias measures differ in how well they align with downstream allocational bias, suggesting that fairness assessments should account for how bias propagates through the broader machine learning pipeline rather than relying on isolated metrics alone [16]. Understanding these interactions requires systematic evaluation of multi-level mitigation strategies across diverse datasets and modeling contexts.

To address these challenges, this paper presents *FairSelect*, a modular toolkit designed to systematically evaluate combinations of fairness techniques across pre-processing, in-processing, and post-processing stages of machine learning pipelines while explicitly supporting intersectional fairness assessment. The toolkit enables users to define predictive models, selected protected characteristics, and apply multiple fairness interventions both individually and concurrently. It also provides quantitative comparisons of the fairness and performance outcomes across demographic and intersectional subgroups for a baseline model, single-method model, and multi-level configuration model. This work therefore extends prior multi-level fairness evaluation efforts to support systematic intersectional analysis and fairness-performance tradeoffs in real-world contexts.

## 2 RELATED WORK.

A growing body of research has explored strategies for mitigating bias across multiple stages of the machine learning lifecycle [8,9,10]. These multi-level fairness approaches, combining pre-processing, in-processing, or post-processing techniques, aim to address multiple sources of bias simultaneously. Several studies in non-clinical domains have demonstrated the feasibility of applying sequential fairness interventions to improve equity while maintaining predictive performance. For example, multi-stage bias mitigation pipelines have been explored in domains such as credit risk assessment and criminal justice decision-making, where combinations of reweighting, fairness constrained optimization, and post-hoc thresholding have been used to reduce disparities across demographic groups [11,12]. These efforts demonstrate fairness interventions may produce complementary effects when applied across different phases of the modeling pipeline. However, such approaches are often developed within a single application context and rarely include a systematic evaluation of multiple combinations of techniques.

Within healthcare settings, research into multi-level fairness strategies remains comparatively limited. Many clinical studies have applied individual fairness mitigation techniques to predictive models, including sampling-based approaches to address representation imbalance and optimization-based methods to constrain subgroup disparities. Fewer studies have explored combined approaches. Of note is study evaluating a model that predicts 2-year stroke risk in Atrial Fibrillation patients using a single combined in-processing and post-processing technique [17]. Another relevant study compared single and multi-level fairness implementations across breast cancer diagnosis, stroke prediction, and Alzheimer’s detection tasks and found that multi-level approaches did not consistently yield superior fairness-accuracy tradeoffs [8]. As a result, the overall general behavior of multi-level fairness strategies across clinical contexts remains poorly characterized.

In addition to multi-level fairness research, there has been an increasing recognition of the importance of intersectional fairness which evaluates disparities across combinations of demographic attributes as opposed to a single demographic axis. Early foundational work introduced formal definitions of intersectional fairness guarantees, including differential fairness, which constrains disparities across all subgroup combinations simultaneously [18]. Subsequent studies have demonstrated the importance of intersectional auditing across domains, revealing disparities that were not detected using traditional single-axis fairness metrics [13,14,15]. These findings reinforce the importance of evaluating fairness across intersecting identities, particularly in healthcare environments characterized by heterogeneous patient populations.

A substantial ecosystem of software tools has been developed to support fairness evaluation and bias mitigation in machine learning. Widely used Python libraries include *AI Fairness 360* (*AIF360*) [19], *Fairlearn* [20], *Aequitas* [21], and *Fairkit-learn* [22], each of which provide implementations of common fairness metrics and selected mitigation algorithms. Additional frameworks such as *TensorFlow Fairness Indicators* [23], *FairLogue* [24], *FairCare* [25], and *FairLens* [26] focus primarily on fairness evaluation and do not offer systemic evaluations of model architectures or techniques. Within the R ecosystem, packages such as *fairmodels* [27] and *fairness* [28] offer complementary functionality for bias assessment and post-hoc fairness analysis. While these tools provide important infrastructure for identifying and mitigating bias, they are generally designed to support individual mitigation strategies rather than structured evaluation of multiple techniques applied concurrently. Furthermore, support for intersectional subgroup evaluation varies widely across implementations.

In addition to methodological and software development efforts, several studies have examined the underlying sources of bias in machine learning systems and the limitations of traditional fairness evaluation approaches. Comprehensive reviews have documented common sources of bias arising from data collection practices, measurement processes, and modeling assumptions, highlighting the importance of understanding sociotechnical drivers of inequity in predictive modeling [2,29,30,31]. These reviews emphasize that single-axis fairness evaluations may overlook complex disparities arising from interactions between demographic and contextual factors, reinforcing the need for more comprehensive evaluation strategies.

Taken together, the existing literature demonstrates growing recognition of the need for both multi-level and intersectional fairness approaches. However, current methods are typically evaluated independently, applied in limited combinations, or implemented within narrowly scoped software environments. There remains a lack of systematic frameworks capable of exploring the combinatorial space of fairness mitigation strategies while supporting intersectional evaluation across diverse clinical contexts. Addressing this gap requires tools that enable reproducible and scalable assessment of fairness interventions across multiple modeling stages, motivating the development of the approach presented in this work.

## 3 METHODS

*FairSelect* was developed as a modular Python-based framework designed to evaluate the effects of fairness mitigation strategies applied across multiple stages of the machine learning lifecycle. The framework enables systematic evaluation of combinations of pre-processing, in-processing, and post-processing fairness techniques applied individually or concurrently within a unified model pipeline. The framework follows a structured execution sequence consisting of three primary phases: (1) pre-processing interventions, (2) model training with in-processing mitigation, and (3) post-processing prediction refinement. Each iteration of the framework begins with a baseline model trained without fairness interventions, followed by independent applications of each selected mitigation technique, and concludes with a combined multi-level configuration with all selected techniques.

Across the three execution phases, a total of twelve fairness mitigation techniques were implemented within the *FairSelect* toolkit, including three pre-processing techniques, five in-processing techniques, and four post-processing techniques. Pre-processing methods modify the training dataset to address representation imbalance and label bias prior to model fitting, in-processing methods incorporate fairness-aware learning strategies during model training, and post-processing methods adjust predicted probabilities or classification thresholds to improve subgroup-level outcomes without modifying the underlying model. These techniques were designed to operate both independently and in combination within the sequential pipeline described above. Detailed descriptions of each mitigation technique are provided in Table 1.

Table 1: List of pre, in, and post-processing techniques made available within FairSelect

| **Technique** | **Primary Bias Addressed** | **Method Summary** | **Primary Effect** |
|---|---|---|---|
| | | **Pre-processing Techniques** | |
| Outcome-Group Reweighting [33] | Outcome-group dependence bias | Assigns instance weights based on joint distribution of outcomes and protected attributes | Reduces learned associations between group membership and outcomes |
| SMOTE [34] | Representation imbalance | Generates synthetic minority samples using interpolation between existing observations | Expands minority representation |
| Local Massaging [33] | Label bias | Selectively modifies labels near decision boundaries within demographic groups | Equalizes positive outcome rates |
| | | **In-processing Techniques** | |
| Exponentiated Gradient Reduction [35] | Predictive disparity across groups | Iteratively updates model weights under fairness constraints | Enforces parity in TPR/FPR across groups |
| Compositional Per-Group Modeling [41] | Subgroup heterogeneity | Trains separate classifiers for each demographic subgroup | Captures group-specific feature relationships |
| Group-Balanced Ensemble [38] | Representation instability | Uses balanced bootstrap sampling across groups | Improves subgroup prediction stability |
| Prejudice Remover Regularization [36] | Prediction dependence on protected attributes | Adds fairness-aware penalties to model objective | Encourages independence between predictions and protected attributes |
| Multicalibration Isotonic Regression [37] | Probability miscalibration | Fits subgroup-specific isotonic regression models | Improves subgroup calibration consistency |
| | | **Post-processing Techniques** | |
| Multiaccuracy Boosting [39] | Residual subgroup error | Iteratively corrects subgroup prediction residuals | Reduces systematic subgroup prediction bias |
| Group-Specific Youden Thresholding [17] | Classification imbalance | Computes subgroup-specific thresholds maximizing Youden's J statistic | Balances sensitivity and specificity across groups |

| Reject-Option Threshold Adjustment [40] | Decision boundary disparity | Adjusts classification thresholds based on subgroup performance | Improves classification parity |
|---|---|---|---|

[a] Descriptions of each bias mitigation technique made available in FairSelect along with description of the type of bias they address. Additional context and implementation details can be found in the source code files

### 3.1 Model Evaluation and Metrics

Model performance and fairness were evaluated using a consistent set of predictive and subgroup-specific metrics. Performance metrics included: Area Under the Receiver Operating Characteristic Curve (AUROC), Accuracy (ACC), F1 Score, Brier Score, and Expected Calibration Error (ECE). Fairness metrics included: Demographic Parity (DP), Equalized Odds (EO), Equal Opportunity Difference (EOD), True Positive Rate (TPR), False Positive Rate (FPR), and Predictive Positive Rate (PPR). These metrics are computed across both individual demographic attributes and intersectional subgroup combinations.

### 3.2 Supported Model Types

*FairSelect* is designed to support multiple model architectures to ensure compatibility with diverse modeling workflows. Users may select from seven commonly used supervised learning algorithms representing both linear and nonlinear model classes. These supported model types include Logistic Regression, Random Forest, Decision Tree, Neural Network, Support Vector Machine (SVM), Extreme Gradient Boosting (XGBoost), and Light Gradient Boosting Machine (LightGBM). All models are trained using a shared preprocessing pipeline to ensure consistent feature transformation across experiments. Model hyperparameters may be specified by the user, enabling customization for domain-specific tasks while preserving the fairness evaluation workflow.

### 3.3 Synthetic Data Generation

To evaluate the behavior of the various fairness techniques under controlled conditions, a series of synthetic datasets were generated to simulate specific types of bias (i.e., representation bias, aggregation bias, calibration bias, etc.) and data imbalance commonly encountered in clinical prediction tasks. These datasets were designed as mechanistically focused stress tests, where each dataset intentionally introduces a structured bias corresponding to the expected correction mechanism of a specific fairness intervention. Synthetic datasets were generated using a template-based simulation framework derived from a real-world clinical prediction context inspired by prediction of glaucoma intervention [3]. Feature distributions were initialized using empirical summary statistics from a template dataset, including demographic variables (Race, Gender), clinical features (e.g., blood pressure measurements, medication history), and derived risk scores. Continuous variables were sampled using Gaussian-based distributions with subgroup-specific variation, while binary features were generated using logistic probability models conditioned on latent severity variables. Outcome labels were generated using logistic transformations of risk signals, with additional group-specific distortions applied to simulate different forms of bias or risk scenarios. Each synthetic dataset was produced using a reproducible randomized seed and exported as a CSV file. A manifest file recorded the parameters used to generate each dataset, including group imbalance factors, feature shifts, label distortions, and calibration modifications. This structured generation approach enables direct mapping between bias mechanisms and fairness strategies, allowing evaluation of whether specific techniques corrected the targeted bias pattern.

A baseline synthetic dataset was first constructed to provide a reference condition representing realistic subgroup variability without a dominant bias mechanism. Additional synthetic datasets were then generated as controlled stress-test scenarios, each designed to isolate a specific source or manifestation of bias relevant to fairness evaluation. While not completely exhaustive, these scenarios provide interpretable validation conditions for assessing whether fairness interventions produced effects consistent with their intended mechanisms.

Several datasets targeted data-level sources of bias. Representation bias and class imbalance were simulated by creating severe subgroup imbalance and rare subgroup-positive outcome combinations to allow for evaluation of synthetic minority over-sampling technique (SMOTE) to improve learning for underrepresented outcome-subgroup strata. Outcome-group dependence was introduced to simulate historical bias or sampling-related associations between protected attributes and outcome labels, providing a setting for evaluating reweighting approaches that aim to reduce the influence of protected attribute imbalance in model training. Localized label bias was simulated by introducing label distortions near the classification boundary, allowing evaluation of local massaging approaches designed to correct potentially biased labels in regions where the classification decisions are highly uncertain.

Additional datasets were generated to target model-level and subgroup structure related bias mechanisms. Explicit dependence between protected characteristics and predicted outcomes was introduced to evaluate fairness-aware optimization strategies, such as prejudice-removal methods, that penalize reliance on protected group membership during model learning. Subgroup specific feature-outcome relationships were simulated to represent aggregation bias, in which a single global model may inadequately capture heterogenous relationships across demographic subgroups. These scenarios provided validation conditions for compositional or subgroup-aware modeling approaches that are intended to improve performance and fairness when predictive relationships differ between groups.

A final set of datasets was constructed to evaluate post-processing strategies that modify inputs, scores, or predictions after model training. Group-specific measurement shifts were introduced into feature distributions to simulate measurement bias, providing a controlled setting for evaluating standardization approaches. Residual prediction errors correlated with subgroup membership were embedded to assess whether multiaccuracy boosting could reduce systematic subgroup-specific error patterns not corrected by the base model. Group-specific score distributions were modified to produce different optimal decision thresholds across subgroups, enabling evaluation of group-wise Youden threshold optimization. Finally, predicted probabilities were concentrated near the decision boundary to simulate threshold-sensitive disparities, allowing evaluation of reject-option threshold shifting.

Together, these datasets form a controlled validation suite that enables systematic testing of individual fairness interventions under controlled bias mechanisms while also supporting evaluation of combined multi-level mitigation strategies across diverse simulated fairness conditions.

### 3.4 Evaluation Approach

Baseline models were first trained using supported model architecture without fairness interventions. Subsequently, individual fairness techniques were applied independently, followed by evaluation of multi-level combinations incorporating pre-processing, in-processing, and post-processing methods.

For each model configuration, predictive performance was evaluated using area under the receiver operating characteristic curve (AUROC) and classification accuracy (ACC). Fairness was evaluated using demographic parity difference (DP_diff) and equalized odds difference (EO_diff), computed across demographic subgroups defined by race and gender. Lower values of DP_diff and EO_diff indicate improved subgroup equity. Changes in fairness and performance metrics were calculated relative to baseline values for each model. Single-method experiments were used to evaluate the independent impact of individual mitigation techniques, while combined-method experiments were used to evaluate interactions between techniques applied across multiple stages of the modeling pipeline.

## 4 COHORT

To evaluate the performance of *FairSelect* in a real-world clinical dataset, a replication-based study was conducted using a previously published machine learning framework for predicting 2-year stroke risk among patients with atrial fibrillation (AF) [17]. The original study demonstrated fairness-aware model tuning and subgroup equity using structured electronic health record (EHR) data from the *All of Us Research Program* [32].

### 4.1 Cohort Selection

Participants were defined as adults aged 18 years or older with a documented diagnosis of atrial fibrillation. The outcome of interest was the occurrence of a stroke within two years following an index date defined as the end of the patient's first qualifying visit. Patients were classified as cases if a stroke occurred within the outcome window and as controls if no stroke occurred, and follow-up data was available beyond the outcome period. This outcome definition aligns with previously described cohort construction strategies for stroke prediction using longitudinal EHR data [17].

To evaluate subgroup generalizability under demographic distribution shifts, the development cohort included participants from states with relatively lower proportions of Black/African American AF populations (WI, MN, AZ, MA, PA, and TX) while the testing cohort included participants from states with higher proportions (FL, MI, GA, IL, and AL). Index dates for the development cohort ranged from January 1, 2018, to December 31, 2019, while testing cohort index dates ranges from January 1, 2020, to December 31, 2021, enabling both geographic and temporal validation of model performance. The final cohort included a total of 11,160 participants, with 8,777 assigned to the development set and 2,383 assigned to the testing set. The overall demographic distribution was predominantly White (86.67%) with Black/African American participants comprising the remaining portion. The gender distribution was approximately balanced, with males representing 52.64% of the total population. The overall incidence of stroke within two years was 8.75% with the development and testing cohorts demonstrating stroke rates of 7.11% and 14.81%, respectively. Demographic characteristics of the cohort are summarized in Table 2.

Table 2: Demographic Characteristics for 2-year Stroke Risk Prediction in AF participants

| Characteristic | Category | Development Set | Testing Set |
|---|---|---|---|
| States Included | — | WI, MN, AZ, MA, PA, TX | FL, MI, GA, IL, AL |
| Total Cohort Size | — | 8,777 | 2,383 |
| Age (years) | Mean (std.) | 67.45 (10.75) | 67.48 (10.97) |
| Race | White | 7,968 (90.78%) | 1,704 (71.51%) |
| | Black or African American | 809 (9.22%) | 679 (28.49%) |
| Gender | Male | 4,642 (52.89%) | 1,233 (51.74%) |
| | Female | 4,135 (47.11%) | 1,150 (48.26%) |

[a] Demographic distributions for each of the development and testing set including Race, Gender, Geographic location, and Age

### 4.2 Data Extraction

Clinical data was obtained from structured EHR data harmonized according to the Observational Medical Outcomes Partnership (OMOP) Common Data Model. Data extraction followed previously described procedures for stroke risk prediction among AF patients within the *All of Us* dataset [17]. For each patient record, the observation window included all available clinical records from the first recorded encounter until the defined index date. The outcome window extended two years following the index date. Clinical conditions were identified using standardized diagnosis codes, while demographic variables including age, race, and gender were extracted from patient-level records. Data was partitioned into development and testing cohorts according to predefined geographic and temporal criteria. Within the development cohort, a subset of patients from PA and TX was used as a tuning population due to higher representation of Black/African American participants, accounting for approximately 30% of the development cohort. This tuning strategy supported subgroup-aware hyperparameter optimization and fairness calibration procedures.

### 4.3 Feature Choices

Predictive features were constructed using clinical variables derived from established stroke risk scoring frameworks, including predictors used in the CHADS2 and CHA2DS2-VASc scores. These features included demographic characteristics, comorbid conditions, and clinical indicators associated with stroke risk. Demographic features included age, race, and gender. Clinical risk factors included conditions such as congestive heart failure, hypertension, diabetes mellitus, vascular disease, and prior stroke history. Additional clinical measurements included physiological and laboratory variables known to be associated with stroke risk, including systolic and diastolic blood pressure, creatinine level, body mass index, and low-density lipoprotein cholesterol level. Feature construction procedures followed previously described ontology-based grouping strategies to aggregate diagnosis and treatment information into clinically meaningful predictors [17]. These included grouping ICD-10 diagnosis codes into higher-level diagnostic categories, aggregating medication codes into pharmacologic classes, and summarizing procedural codes into clinically interpretable groupings. After applying the ontology roll-up-based grouping, a total of 385 clinical features were included in the model.

## 5 RESULTS FROM SYNTHETIC EXPERIMENTS

Across all synthetic scenarios, baseline models demonstrated measurable disparities consistent with the intended bias structures introduced during dataset generation. Targeted fairness techniques produced changes in fairness metrics consistent with their expected correction mechanisms, supporting the validity of the implementation and execution workflow. Individual fairness interventions produced modest reductions in subgroup disparities with minimal impact on predictive performance.

1. Across single-method configurations, EO_diff (equalized odds difference) improved by an average of 0.0175 and DP_diff (demographic parity difference) by 0.0153, while mean AUROC decreased by 0.0014 and mean accuracy decreased by 0.0036. Multi-level mitigation strategies produced larger average fairness improvements, with mean EO_diff and DP_diff improvements of 0.0331 and 0.0285, respectively.
2. Across multi-level configurations, mean AUROC decreased by 0.0125 and mean accuracy decreased by 0.0206, compared to smaller reductions observed in single-method experiments ($\Delta$AUROC = −0.0014, $\Delta$ACC = −0.0036). This pattern indicates that stronger fairness gains were often associated with modest reductions in predictive performance, consistent with expected fairness–utility tradeoff dynamics observed in prior fairness literature.

Importantly, improvements were not uniform across all combinations. While many multi-level configurations produced larger fairness reductions than individual techniques, some combinations produced limited additional benefit or, in certain cases, resulted in reduced fairness gains relative to single-method configurations. This variability is reflected in the wide observed EO_diff range for combined strategies (−0.234 to 0.938) compared to single-method strategies (−0.146 to 0.583). These findings demonstrate that fairness interventions interact in non-additive ways and reinforce the importance of systematically evaluating combinations rather than assuming independent or cumulative effects.

## 6 TWO-YEAR STROKE RISK IN AF PARTICIPANTS RESULTS

### 6.1 Baseline Performance

Baseline models showed strong predictive performance but substantial variability in fairness across model architectures (Table 3). AUROC ranged from 0.653 to 0.826, and accuracy ranged from 0.867 to 0.913. LightGBM achieved the highest baseline AUROC, while random forest showed the most favorable baseline fairness profile. In contrast, the decision tree model exhibited the largest baseline disparities.

Table 3: Baseline model performance and fairness

| Model | AUROC | ACC | DP_diff | EO_diff |
|---|---|---|---|---|
| LightGBM | 0.826 | 0.913 | 0.087 | 0.360 |
| XGBoost | 0.824 | 0.913 | 0.099 | 0.500 |
| Random Forest | 0.824 | 0.907 | 0.032 | 0.063 |
| SVM | 0.798 | 0.910 | 0.037 | 0.174 |
| Logistic Regression | 0.773 | 0.901 | 0.154 | 0.329 |
| Neural Network | 0.666 | 0.900 | 0.102 | 0.269 |
| Decision Tree | 0.653 | 0.867 | 0.308 | 1.000 |

[a] Performance and Fairness Metrics for each of the baseline model types on the two-year stroke risk in AF participants cohort

### 6.2 Single-Method Performance

Across all single-method configurations, fairness improvements were inconsistent. Only 22.3% of single method runs reduced EO_diff, while 27.4% reduced DP_diff relative to baseline models. On average, single-method interventions produced a mean increase in EO_diff of 0.0720 and a mean increase in DP_diff of 0.0608, indicating that many individual fairness techniques did not improve intersectional subgroup equity in this clinical prediction task. Predictive performance was largely preserved across single-method experiments, with a mean decrease in AUROC of 0.011 and mean accuracy decreasing by 0.015. The most favorable single-method results were observed primarily for random forest and SVM models (Table 4). In particular, random forest with ensemble in-processing achieved the best overall balance of fairness and utility, while random forest with multiaccuracy boosting produced the lowest EO_diff among single-method runs with essentially unchanged AUROC. SVM with reweighting also yielded a substantial reduction in EO_diff with minimal performance loss.

Table 4: Selected best single-method configurations

| Model | Technique | AUROC (Δ) | ACC (Δ) | DP_diff (Δ) |
|---|---|---|---|---|
| Random Forest | In: Ensemble (K=5) | 0.809 (-0.015) | 0.909 (0.002) | 0.012 (-0.02) |

| Random Forest | Post: Multiaccuracy Boost | 0.825 (0.001) | 0.906 (0) | 0.032 (0) |
|---|---|---|---|---|
| SVM | Pre: Reweight | 0.799 (0.001) | 0.908 (-0.002) | 0.037 (0) |
| Logistic Regression | Pre: Local Massaging | 0.748 (0.238) | 0.901 (0) | 0.154 (0) |
| Neural Network | Post: Multiaccuracy Boost | 0.674 (0.008) | 0.902 (0.002) | 0.094 (-0.008) |

[a] Performance and Fairness Metrics for five of the best performing models using a single fairness technique application

### 6.3 Multi-Level Combined Strategy Performance

Multi-level mitigation strategies produced the strongest overall fairness improvements but also demonstrated substantial variability depending on the combination of techniques used. Across all multi-level approaches, only 26.2% reduced EO_diff and 26.2% reduced DP_diff relative to baseline models. Mean fairness changes across all combinations were +0.1627 for EO_diff and +0.1589 for DP_diff, indicating that many combinations were ineffective or counterproductive. Despite this variability, several combined configurations produced meaningful fairness gains while maintaining competitive predictive performance (Table 5). Combined methods also exhibited larger average performance reductions than single-method approaches, with mean AUROC decreasing by 0.0480 and mean accuracy decreasing by 0.0262. The strongest overall combined fairness profile was observed for random forest with reweighting and ensemble in-processing, with and without reject-option shift. Decision Tree combinations showed the largest absolute reductions in subgroup disparity, particularly for EO_diff, although these gains should be interpreted in the context of the model's poor baseline fairness and lower initial predictive performance. SVM with reweighting and compositional per-group modeling also substantially reduced EO_diff with minimal change in accuracy, while LightGBM achieved modest fairness gains under local massaging, compositional modeling, and multiaccuracy boost. No Neural Network configuration was retained among the best-performing combined methods because no combined method consistently improve both DP_diff and EO_diff.

Table 5: Selected best combined methods configurations

| **Model** | **Combined strategy** | **AUROC (Δ)** | **ACC (Δ)** | **DP_diff (Δ)** | **EO_diff (Δ)** |
|---|---|---|---|---|---|
| Random Forest | Reweight -> Ensemble (K=5) | 0.813 (-0.01) | 0.908 (0.002) | 0.012 (-0.02) | 0.014 (-0.049) |
| Random Forest | Reweight -> Ensemble (K=5) -> Reject-Option Shift | 0.813 (-0.011) | 0.908 (0.002) | 0.006 (-0.026) | 0.018 (-0.045) |
| Decision Tree | Reweight -> Ensemble (K = 5) -> Reject-Option Shift | 0.757 (0.105) | 0.889 (0.022) | 0.112 (-0.196) | 0.308 (-0.692) |

| Decision Tree | Local Massaging -> Ensemble (K=5) -> Multiaccuracy Boost | 0.740 (0.087) | 0.900 (0.033) | 0.096 (-0.212) | 0.348 (-0.652) |
|---|---|---|---|---|---|
| SVM | Reweight -> Compositional per-group | 0.759 (-0.039) | 0.910 (0) | 0.027 (-0.023) | 0.138 (-0.362) |
| LightGBM | Local Massaging -> Compositional per-group -> Multiaccuracy Boost | 0.781 (-0.045) | 0.911 (-0.002) | 0.075 (-0.012) | 0.346 (-0.014) |

[a] Performance and Fairness Metrics for five of the best performing models using a multi-level fairness technique application

## 7 DISCUSSION

This study presents *FairSelect*, a modular framework designed to systematically evaluate fairness mitigation strategies across multiple stages of the machine learning lifecycle. Through both synthetic and real-world clinical evaluation, this work demonstrates that fairness interventions behave in complex and highly context-dependent ways, and that systematic evaluation of mitigation strategies is necessary to identify effective configurations. The synthetic experiments confirmed that the implemented fairness techniques behaved consistently with their intended design under controlled conditions. Targeted bias scenarios produced expected baseline disparities, and individual mitigation methods reduced those disparities when applied to datasets designed to match their theoretical assumptions. Application of multiple, complementary mitigation strategies generally produced larger average fairness improvements than individual methods alone, demonstrating that multi-level techniques can address multiple sources of bias simultaneously.

Results from the real-world clinical task reveal a more complex picture. Although baseline models demonstrated strong predictive performance, they retained substantial subgroup disparities, confirming that predictive accuracy alone does not ensure equitable outcomes. More importantly, fairness interventions were highly variable in their effectiveness: only a minority of single-method runs improved fairness metrics, and many produced no benefit or even worsened subgroup disparities. This finding highlights a critical limitation in current fairness practice in that the assumption that applying any algorithmic fairness method will reliably improve model equity is not always true. In reality, the effectiveness of any intervention depends strongly on the interaction between the method, model architecture, dataset characteristics, and underlying sources of bias.

Multi-level mitigation strategies provided stronger evidence for the value of systematic combination testing. Combined configurations were more likely than single-method configurations to improve both DP_diff and EO_diff, and several combinations achieved large fairness gains while preserving or improving predictive performance. The strongest improvements occurred when the baseline model had substantial fairness gaps, as seen in decision tree and neural network configurations. In these cases, combinations such as reweighting with compositional modeling and reject-option thresholding, or reweighting with multicalibration and reject-option thresholding, reduced intersectional disparities while also improving AUROC and accuracy. These findings suggest that multi-level interventions can sometimes correct bias patterns that also degrade model generalization, making fairness and predictive performance complementary rather than strictly competing objectives.

These results highlight the importance of moving beyond single-method fairness evaluation toward systematic exploration of multi-level fairness strategies. Existing fairness research often evaluates mitigation methods in isolation, frequently under controlled experimental conditions that do not reflect the complexity of real-world deployment environments. These findings demonstrate that such isolated evaluation may provide an incomplete understanding of how fairness interventions behave when integrated into full modeling pipelines. In clinical contexts, where predictive models directly influence patient care decisions, the consequences of unintended fairness failures can be substantial. Systematic evaluation frameworks such as *FairSelect* provide a mechanism to identify effective mitigation strategies prior to deployment, reducing the risk of introducing new disparities during model implementation.

Another important contribution of this work is the demonstration that fairness interventions must be evaluated alongside predictive performance rather than independently. Several fairness methods produced substantial reductions in subgroup disparities but at the cost of large reductions in AUROC, illustrating the well-known fairness–utility tradeoff. However, other combinations achieved substantial fairness gains while maintaining or even improving predictive performance, suggesting that fairness and performance are not always competing objectives. Instead, carefully selected multi-level mitigation strategies may improve model generalization by correcting systematic biases that degrade predictive accuracy.

This work also reinforces the importance of considering intersectional subgroup structure when evaluating fairness interventions. Although many fairness methods are designed to address disparities across single demographic attributes, real-world inequities often arise through overlapping demographic characteristics. The modular design of *FairSelect* enables evaluation across both demographic and intersectional subgroups, allowing developers to identify disparities that may be masked by single-axis analyses. Future work should expand this capability to include additional protected attributes and more granular intersectional definitions to better reflect the complexity of real-world populations.

Despite these contributions, several limitations should be considered. First, the clinical evaluation focused on a single prediction task involving stroke risk among patients with atrial fibrillation. Although this task provides a meaningful validation scenario, fairness intervention behavior may vary across clinical domains with different data distributions and outcome prevalence. Second, while the synthetic datasets were designed to simulate common bias mechanisms, they cannot capture the full complexity of biases present in real-world healthcare systems. Third, the combinatorial space of fairness techniques grows rapidly as additional methods are introduced, making exhaustive evaluation computationally expensive. Future work should explore adaptive search strategies or automated selection methods to identify promising combinations more efficiently. Lastly, the interpretability of the fairness metrics may degrade as the number of intersectional subgroups increases due to low representation in each particular subgroup. The toolkit does offer limited tools to filter out low subgroups, but this may still impact the generalizability of the aggregate fairness metrics to all intersections.

## 8 CONCLUSION

This study demonstrates that fairness mitigation in clinical machine learning is not a one-size-fits-all process. Instead, fairness outcomes depend on complex interactions between modeling choices, dataset characteristics, and intervention strategies. The results presented here provide empirical evidence that systematic, multi-level evaluation frameworks are necessary to identify effective fairness interventions in real-world settings. Through enabling structured exploration and evaluation of fairness techniques across multiple stages of the modeling pipeline, *FairSelect* offers a practical tool for advancing equitable machine learning development in healthcare.